\pdfoutput=1

\documentclass[11pt]{article}

\usepackage[final]{acl}

\usepackage{times}
\usepackage{latexsym}

\usepackage[T1]{fontenc}

\usepackage[utf8]{inputenc}
\usepackage{url}            
\usepackage{booktabs}       
\usepackage{amsfonts}       
\usepackage{nicefrac}       

\usepackage{xcolor}         
\usepackage{graphicx}
\usepackage{caption}
\usepackage{subcaption}

\usepackage{amssymb,amsmath,amsthm}
\usepackage{lipsum} 

\newtheorem{assumption}{Assumption}

\usepackage{microtype}
\usepackage{amsmath} 
\usepackage{algorithm} 
\usepackage{natbib}
\usepackage{newfloat}
\usepackage{algorithm}
\usepackage{algorithmic}
\usepackage{listings}

\usepackage{microtype}

\usepackage{inconsolata}

\usepackage{graphicx}

\title{Bayesian Optimization for Controlled Image Editing via LLMs}

\author{\normalfont
Chengkun Cai\textsuperscript{1}\thanks{Equal contribution.} \quad
Haoliang Liu\textsuperscript{2}\footnotemark[1] \quad
Xu Zhao\textsuperscript{1}\footnotemark[1] \quad
Zhongyu Jiang\textsuperscript{3} \quad
Tianfang Zhang\textsuperscript{4} \\
Zongkai Wu\textsuperscript{5}\thanks{Project Leader.} \quad
John Lee\textsuperscript{1} \quad
Jenq-Neng Hwang\textsuperscript{3} \quad
Lei Li\textsuperscript{3,6}\thanks{Corresponding author: \texttt{lilei@di.ku.dk}} \\
\\
\textsuperscript{1}University of Edinburgh \quad
\textsuperscript{2}University of Manchester \quad
\textsuperscript{3}University of Washington \\
\textsuperscript{4}Tsinghua University \quad
\textsuperscript{5}Skai Intelligence \quad
\textsuperscript{6}University of Copenhagen
}

\begin{document}
\maketitle
\begin{abstract}
In the rapidly evolving field of image generation, achieving precise control over generated content and maintaining semantic consistency remains a crucial limitation, particularly concerning grounding techniques and the necessity for model fine-tuning. To address these challenges, we propose BayesGenie, an off-the-shelf approach that integrates Large Language Models (LLMs) with Bayesian Optimization to facilitate precise and user-friendly image editing. Our method enables users to modify images through natural language descriptions without manual area marking, while preserving the original image's semantic integrity. Unlike existing techniques that require extensive pre-training or fine-tuning, our approach demonstrates remarkable adaptability across various LLMs through its model-agnostic design. BayesGenie employs an adapted Bayesian optimization strategy to automatically refine the inference process parameters, achieving high-precision image editing with minimal user intervention. Through extensive experiments across diverse scenarios, we demonstrate that our framework outperforms existing methods in both editing accuracy and semantic preservation, as validated using different LLMs including Claude3 and GPT-4.

\end{abstract}
\section{Introduction}

In the rapidly evolving field of visual content manipulation, image editing has gained significant attention due to its practical applications across various domains. Unlike traditional image generation models such as Stable Diffusion~\cite{rombach2022high} and DALL-E 3~\cite{ramesh2022hierarchical}, our work is specifically focused on improving control in the image editing process. Recent advancements in controllable synthesis, such as those by Hertz et al. \cite{hertz2022prompt} and Brooks et al. \cite{brooks2023instructpix2pix}, have introduced methods to fine-tune and guide transformations in existing images rather than generating new ones from scratch.

\begin{figure}[t]
    \centering
    \begin{subfigure}[b]{0.23\textwidth}
        \includegraphics[width=\textwidth,height=2.5cm]{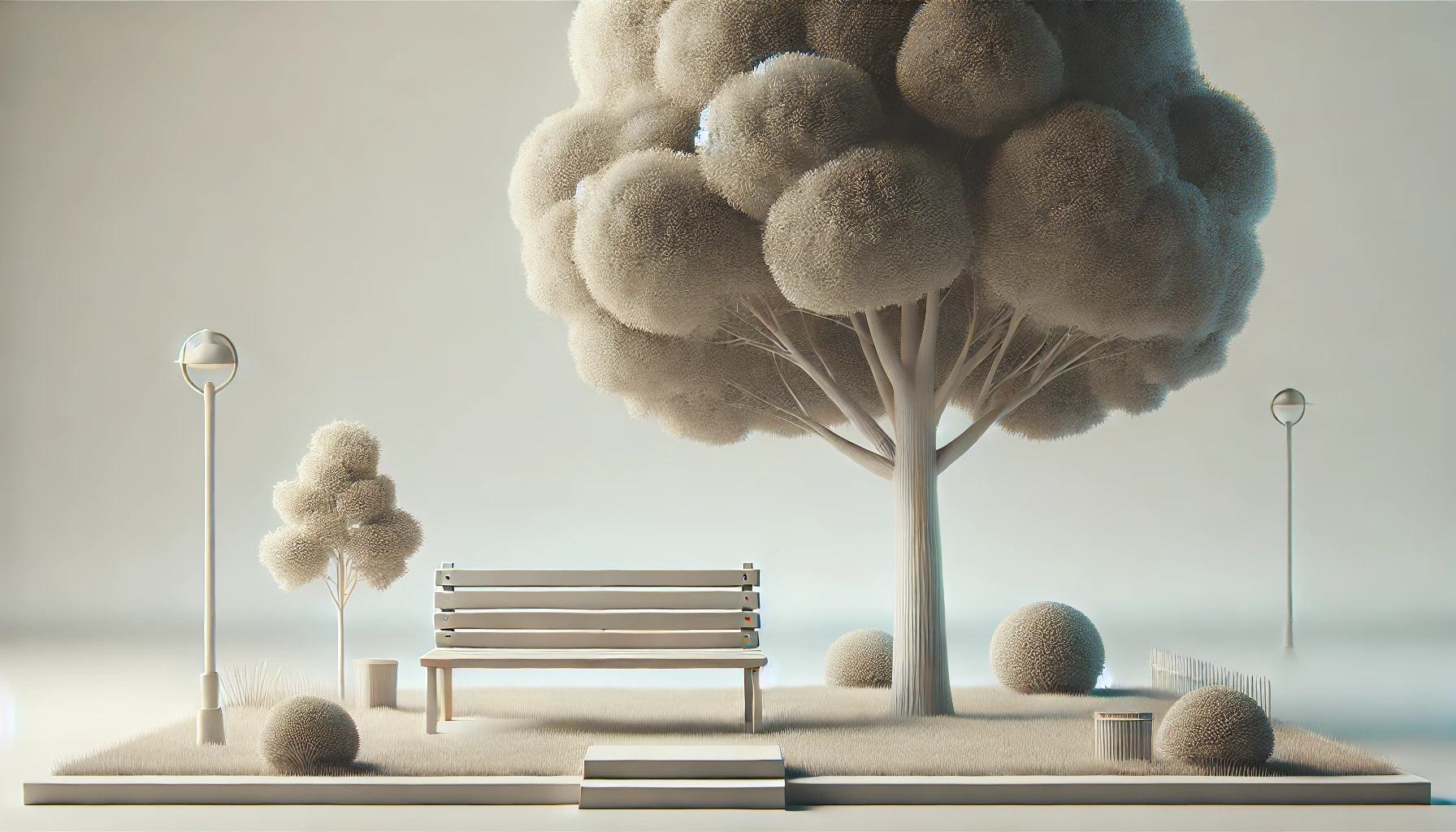}
        \caption{Original image.}
        \label{fig:first}
    \end{subfigure}
    \begin{subfigure}[b]{0.23\textwidth}
        \includegraphics[width=\textwidth,height=2.5cm]{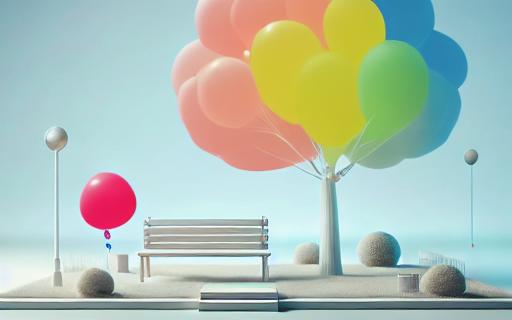}
        \caption{Edited image.}
        \label{fig:second}
    \end{subfigure}

     \caption{Practical application results in design scenarios: "replace the tree next to the bench with balloons"}
    \label{fig:1}
\end{figure}

Controllable synthesis in generation technology~\cite{guan2025learning,yao2024car,hertz2022prompt,brooks2023instructpix2pix,patashnik2021styleclip,jiang2024back} has recently attracted significant attention due to its expanded range of applications. Models such as Pix2Pix and CycleGAN have demonstrated the ability to transform images from one domain to another, effectively applying controllable synthesis to tasks like style transfer and image enhancement \cite{isola2017image,zhu2017unpaired}. Recent advances like ZONE have enabled instruction-driven modifications without pre-defined training samples \cite{li2024zone}, while new frameworks have emerged for intuitive and localized image editing by manipulating internal attention mechanisms \cite{brooks2023instructpix2pix}. This represents a shift towards more granular control over AI-generated content, enabling precise, region-specific alterations without additional input masks.

However, despite these advancements, existing methods face several critical challenges. First, most state-of-the-art local editing methods heavily rely on mask priors to constrain the editing regions—either through manual input or derived from attention map analysis and semantic segmentation—making them less accessible for non-expert users. Additionally, methods like cross-attention control and diffusion models often encounter challenges in fine-tuning model parameters to align with user requirements, resulting in a disparity between desired and actual outputs. These issues are particularly pronounced in applications that demand detailed modifications based on user instructions. With the LLMs-Driven adaptation, recent various applications~\cite{cai2024t,li2024human,shi2025explaining,liu2024graph,shi2024chops,cai2024role} have advanced developed.

To address these challenges, we propose BayesGenie: a novel framework that achieves precise localized editing without any form of mask guidance. Our approach uniquely combines the semantic understanding capabilities of LLMs with the parameter optimization power of Bayesian methods~\cite{openai2023,snoek2012practical}. BayesGenie leverages LLMs to generate detailed prompts from user requirements, which then guide a Stable Diffusion model to modify images accurately. The framework employs Bayesian Optimization to systematically explore the parameter space, particularly the image and text Classifier Free Guidance (CFG) weights, to maximize output quality.

Our method provides an end-to-end solution, where users are only required to provide a textual description, eliminating the need for manual selection or marking of specific image regions. This approach streamlines user interaction, enhancing intuitiveness and accessibility. Moreover, our method operates without pre-training or fine-tuning on specific datasets, instead leveraging the capabilities of multiple multimodal LLMs to produce high-quality outcomes. 

Our experimental results demonstrate that the integration of LLMs and Bayesian Optimization enables more intuitive and accurate image editing. As shown in Figure \ref{fig:1}, our method can effectively implement specific modifications while maintaining the scene's overall coherence and aesthetic integrity.

In summary, our contributions are:

\begin{itemize}
    \item We propose BayesGenie, a novel image editing framework that enables precise localized modifications without the need for manual region annotations or mask generation. The framework adopts a model-agnostic design that leverages large language models (LLMs) for spatial reasoning and semantic interpretation, allowing for accurate and context-aware image editing based solely on natural language instructions. Distinctively, BayesGenie is both training-free and mask-free, which endows it with strong generalization capabilities and high adaptability. This design choice allows BayesGenie to be seamlessly integrated with a wide range of existing image editing backbones—regardless of whether the underlying models are mask-based, fine-tuned, or operate in a zero-shot manner.
    
    \item We introduce an automated parameter optimization system based on Bayesian Optimization that eliminates the need for manual parameter tuning or pre-training. This system automatically discovers optimal editing parameters through iterative refinement, independent of the model training, making our framework immediately deployable across different scenarios without requiring specialized adjustments.
    
    \item Through extensive experiments, we demonstrate that our off-the-shelf framework achieves superior performance and broad adaptability across various editing scenarios. The framework's effectiveness has been validated with different multimodal LLMs, showcasing its versatility and robustness while maintaining both local precision and global consistency.
\end{itemize}

\subsection{Related Work}
\paragraph{Image-to-Image Generation Models}
Image-to-image translation models have become increasingly significant in the field of computer vision. Generative Adversarial Networks (GANs) and auto-regressive models have been pivotal, with notable architectures like Instruct Pix2Pix, CycleGAN, and PixelCNN demonstrating impressive results~\cite{isola2017image,zhu2017unpaired,van2016pixel}. Diffusion models, such as SR3 and ADM, have emerged as powerful alternatives, offering superior quality and diversity in image generation tasks by progressively refining noisy images to high-quality outputs~\cite{saharia2022image,dhariwal2021diffusion}. 

 The Instruct Pix2Pix framework represents a significant advancement in the field of image editing~\cite{brooks2023instructpix2pix}. This model has been widely used in various image-to-image generation tasks, such as converting hand-drawn sketches into photographs~\cite{m2022transfer}, transforming abstract maps into realistic map images~\cite{li2024mapping} and de-noising images taken in harsh environments for crowd counting~\cite{khan2023crowd}. Instruct Pix2Pix employs Classifier-Free Guidance (CFG) for both image and text conditions, adjusting the weights of these inputs to control the generated output. It enables users to use natural language instructions for image editing, leveraging the model's ability to implement detailed modifications. The system manipulates the internal attention mechanisms of generative models, offering precise alterations without the need for additional input masks. Innovations such as DALLE-3 and CLIP integrate multi-modal learning, leveraging large-scale text and image datasets to enhance contextual understanding and generation capabilities~\cite{betker2023improving,radford2021learning}. 

However, a common limitation persists across most state-of-the-art approaches: they typically rely on some form of mask guidance to achieve precise local editing. Whether through manual mask annotations~\cite{hertz2022prompt}, attention map analysis~\cite{li2024zone}, or semantic segmentation, the dependence on mask priors creates a barrier for non-expert users and limits the flexibility of these systems. Recent works have begun to overcome this challenge through fully mask-free editing mechanisms. For instance, strong mask-free image editing baselines have emerged, such as MagicBrush~\cite{zhang2023magicbrush} and UltraEdit~\cite{zhao2024ultraedit}, which enable high-quality, instruction-driven editing without relying on explicit masks.


\paragraph{LLM-assisted Image Generation}
LLMs have significantly advanced numerous NLP tasks through their exceptional generalization capabilities, which have also been effectively harnessed to enhance image-to-image generation processes. Flamingo combines visual information with the multimodal generalization capabilities of LLMs, enabling it to handle new tasks without specific training \cite{alayrac2022flamingo}. LayoutGPT utilizes LLMs to interpret structured diagrams, akin to CSS, enabling the accurate positioning of objects within a generated scene, which allows it to understand spatial relationships and apply them consistently across various layouts \cite{NEURIPS2023_3a7f9e48}. 

\begin{figure*}[ht]
    \centering
    \includegraphics[width=\linewidth]{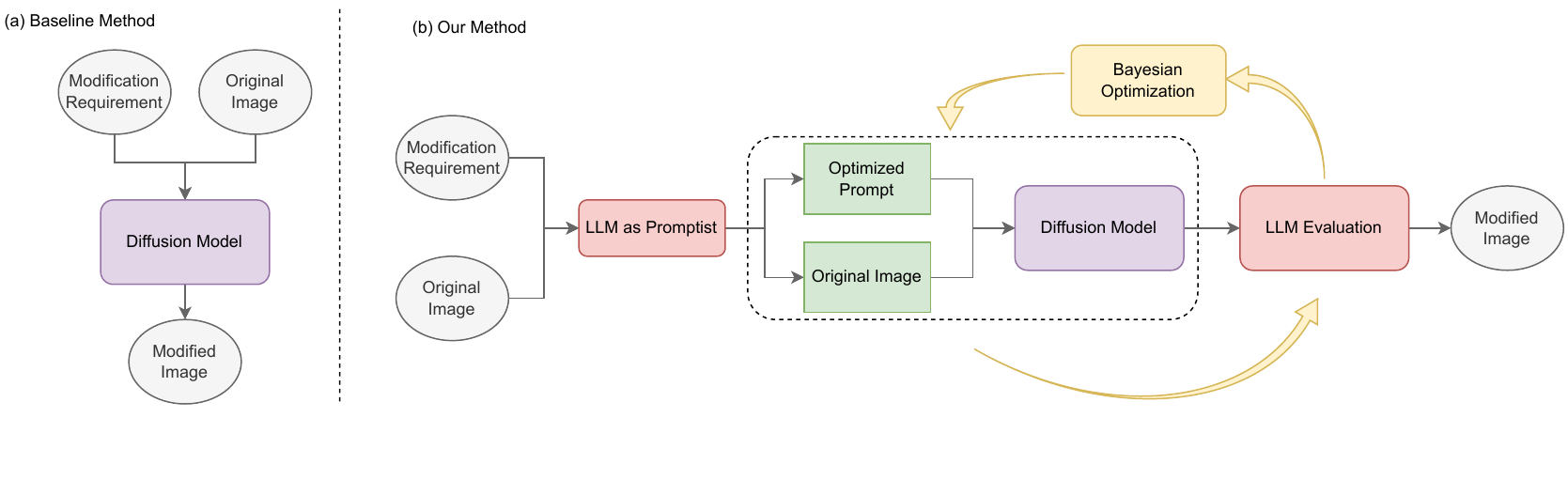}
    \caption{The System Architecture for Fine-Grained Image Control Using LLMs and Bayesian Optimization is detailed herein. Figure (a) illustrates the conventional method for comparison purposes.}
    \label{fig:flow}
\end{figure*}

\paragraph{Bayesian learning}
Black-box functions are frequently encountered across various domains, particularly in the intricate task of parameter tuning within machine learning \cite{JMLR:v25:23-0269}. Bayesian learning \cite{10.1115/1.3653121}, a statistical method, facilitates the inference of model parameters by integrating prior knowledge with the likelihood derived from observed data. In the context of BayesGenie, Bayesian learning is leveraged to optimize the parameters of generative models, thereby enhancing both the quality and diversity of generated images through the minimization of the loss function. Specifically, Bayesian Optimization approximates the objective function by constructing a surrogate model, such as a Gaussian process \cite{Jones1998}, and employs global optimization techniques to identify the optimal model parameters. Currently, Bayesian optimization is widely used for finding the optimal hyperparameters of models\cite{boyar2024latent,aristodemou2025maximizing}.

\section{Methodology}


Our system architecture integrates LLMs and Bayesian optimization for image editing (Figure \ref{fig:flow}). An LLM processes the original image and modification requirements to generate a textual prompt capturing the desired changes. This prompt and the original image are then fed into a diffusion model to generate a modified image.
BayesGenie enhances this process through dynamic prompt refinement and parameter optimization. The LLM evaluates each generated image, scoring it based on requirement satisfaction and providing feedback for improvements. Bayesian optimization then iteratively adjusts key parameters, specifically `text\_cfg\_scale` and `image\_cfg\_scale`, which balance text and image components in the diffusion model. This optimization minimizes the negative LLM score, maximizing alignment between generated images and desired outcomes.

\subsection{Dynamic Prompt Optimization with LLMs}
In our approach, the prompt is dynamically optimized through an iterative optimization process. Initially, the LLM generates a prompt based on the user's modification requirements, which guides the diffusion model to generate a preliminary image. Once the image is generated, the LLM evaluates it by assigning a score based on how well it aligns with the user's specifications and provides feedback on areas needing improvement, such as adding more details or adjusting object positioning. This feedback is then used to refine the prompt for the next iteration. The process repeats, with the refined prompt guiding the generation of a new image, until the desired result is achieved or the maximum number of iterations is reached.

\subsection{Preliminaries}

\begin{assumption}

Consider the existence of a set of optimized guidance scales, denoted as \( s_I^* \) and \( s_T^* \) such that, the generated image \( I_{\text{gen}}(s_I^*, s_T^*) \) produced by the score network \( \tilde{e}_{\theta}(z_t, c_I, c_T) \) satisfies a specific requirement or objective function \( \mathcal{L}(I_{\text{gen}}(s_I^*, s_T^*)) \). Mathematically, this can be expressed as:

\begin{equation}
(s_I^*, s_T^*) = \arg\min_{s_I, s_T} \mathcal{L}\left(I_{\text{gen}}(s_I, s_T)\right)
\end{equation}
where the score network \( \tilde{e}_{\theta}(z_t, c_I, c_T) \) is defined as:
\begin{equation}
\begin{aligned}
    \tilde{e}_{\theta}(z_t, c_I, c_T) &= e_{\theta}(z_t, \emptyset, \emptyset) 
    \\
    &+ s_I \cdot \left( e_{\theta}(z_t, c_I, \emptyset) - e_{\theta}(z_t, \emptyset, \emptyset) \right) 
    \\
    &+ s_T \cdot \left( e_{\theta}(z_t, c_I, c_T) - e_{\theta}(z_t, c_I, \emptyset) \right)
\end{aligned}
\end{equation}

where $z_t$ represents the noisy latent variable at timestep $t$. $\emptyset$ denotes the unconditional input. $e_{\theta}$ represents the score network which estimates the gradient of the noisy latent variable \( z_t \) relative to the clean data given the conditioning inputs \( c_I \) and \( c_T \).
\end{assumption}

This problem formulation stems from the need to address the inherent challenges in fine-grained image modification tasks. While the diffusion models, particularly the Instruct Pix2Pix variant, offer significant control and precision through their diffusion process, the manual adjustment of parameters such as imageCFG and textCFG ($s_I, s_T$).

Given the hypothesis that for any initial image and a sufficiently detailed and comprehensive image generation prompt, there exists a pair of imageCFG and textCFG parameters that ensure the final generated image meets the specified requirements, this problem can be framed as an optimization task. 

In this context, the task becomes one of finding the optimal set of parameters (imageCFG and textCFG) that maximize the quality of the generated image according to the specified criteria. This involves searching through the parameter space to identify the values that produce the best possible image modifications, as defined by the evaluation metrics.

\subsection{Bayesian Optimization}
The optimization process involves the repeated evaluation of the objective function using Bayesian optimization, where each evaluation includes generating a modified image with the current CFG parameters and scoring it through the LLM. The system iteratively samples the parameter space and updates its model of the objective function landscape based on the results of previous evaluations, aiming to find the optimal guidance scales that yield the highest quality edited images.

Specifically, the objective function of Bayesian optimization is 
\begin{equation}
f(\mathbf{s}) = f([s_I, s_T]),
\end{equation}
where $\mathbf{s} = [s_I, s_T]$ represents our two guidance parameters (ImageCFG and TextCFG), and $f(\mathbf{s})$ represents the quality score evaluated by the LLM based on both semantic alignment and image quality.

Bayesian optimization uses a Gaussian Process (GP) to approximate the objective function $f(\mathbf{s})$. The Gaussian Process assumes that all possible function values have a joint Gaussian distribution:
\begin{equation}
f(\mathbf{s}) \sim \mathcal{GP}(\mu(\mathbf{s}), k(\mathbf{s}, \mathbf{s'}))
\end{equation}
where $\mu(\mathbf{s})$ is the mean function and $k(\mathbf{s}, \mathbf{s'})$ is the kernel function that defines the similarity between parameter settings. We employ the Matérn kernel function for its robustness in optimization tasks.

To select the next evaluation point, we use Expected Improvement (EI) as the acquisition function:
\begin{equation}
\text{EI}(\mathbf{s}) = \mathbb{E}[\max(0, f(\mathbf{s}) - f(\mathbf{s}^+))]
\end{equation}
where $\mathbf{s}^+$ represents the current best parameter setting.

The optimization process follows these steps:
\begin{enumerate}
    \item \textbf{Initialization}: Select initial guidance scales $\mathbf{s}_1 = [s_I, s_T]$ and compute their corresponding quality scores $y_i = f(\mathbf{s}_i)$.
    
    \item \textbf{Model Update}: Update the Gaussian Process model using the accumulated observations.
    
    \item \textbf{Select Next Evaluation Point}: Maximize the acquisition function to determine the next parameter setting:
    \begin{equation}
    \mathbf{s}_{n+1} = \arg\max_{\mathbf{s}} \text{EI}(\mathbf{s})
    \end{equation}
    
    \item \textbf{Evaluate Objective Function}: Generate and evaluate an image using $\mathbf{s}_{n+1}$, computing $f(\mathbf{s}_{n+1})$ and adding it to the observation dataset.
    
    \item \textbf{Iterate}: Repeat steps 2-4 until reaching the maximum number of iterations or achieving convergence.
\end{enumerate}

The effectiveness of this approach relies on the smooth relationship between CFG parameters and image quality, where small parameter adjustments typically lead to predictable changes in the output. By systematically exploring the parameter space, our method efficiently discovers optimal guidance scales that balance maintaining original image features with implementing the desired edits.

\subsection{Scoring Evaluation}
In the context of our optimization task, accurate evaluation of the generated images is crucial. While traditional metrics like CLIP scores are commonly used, they fall short in evaluating fine-grained image modifications. To address this limitation, we utilize LLMs for a more nuanced evaluation, leveraging their robust multimodal understanding capabilities.

\begin{figure}[ht]
    \centering
    \includegraphics[width=0.4\textwidth]{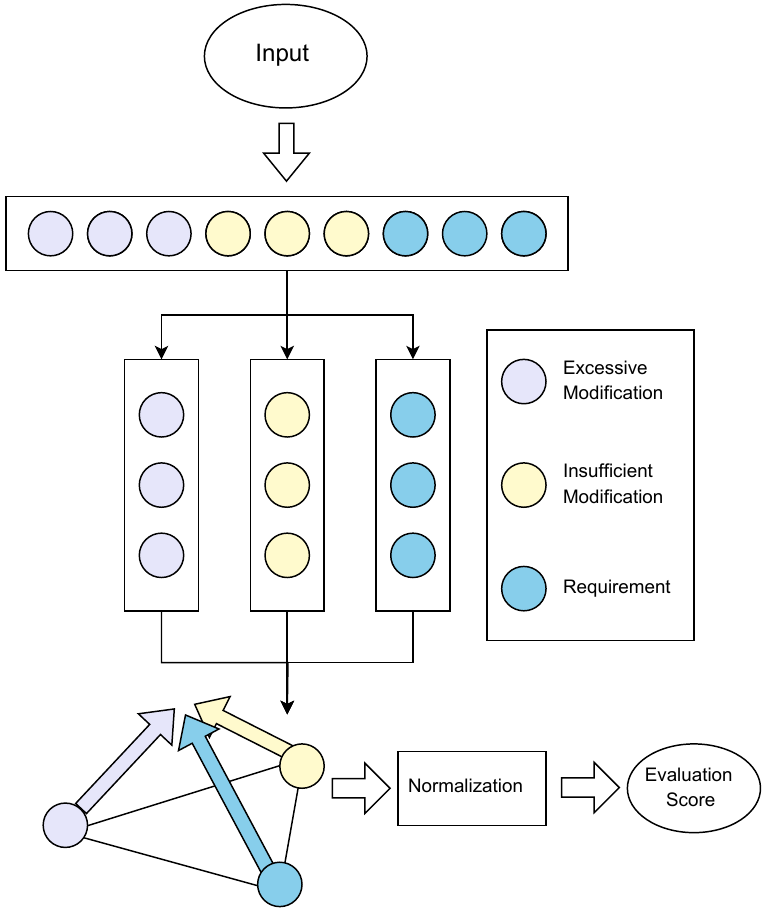}
    \caption{An illustration of the prompt-based LLM evaluation process}
    \label{fig:ptompt}
\end{figure}
Our evaluation is guided by a predefined 0-shot prompt designed to ensure consistency and objectivity (see Supplement for the full prompt). This prompt directs the LLM to integrate three distinct types of constraints—excessive modification, insufficient modification, and compliance with requirements—into a comprehensive, high-dimensional representation of the image's quality. Figure \ref{fig:ptompt} visually outlines how the LLM utilizes this prompt to systematically calculate a score and provide a detailed explanation.


The evaluation process ensures that scores follow a normal distribution, reflecting a balanced judgment across the dataset. Furthermore, the LLM generates concise explanations for each score, enhancing the transparency and interpretability of the evaluation process.

\begin{figure*}[h]
    \centering
    \includegraphics[width=\linewidth]{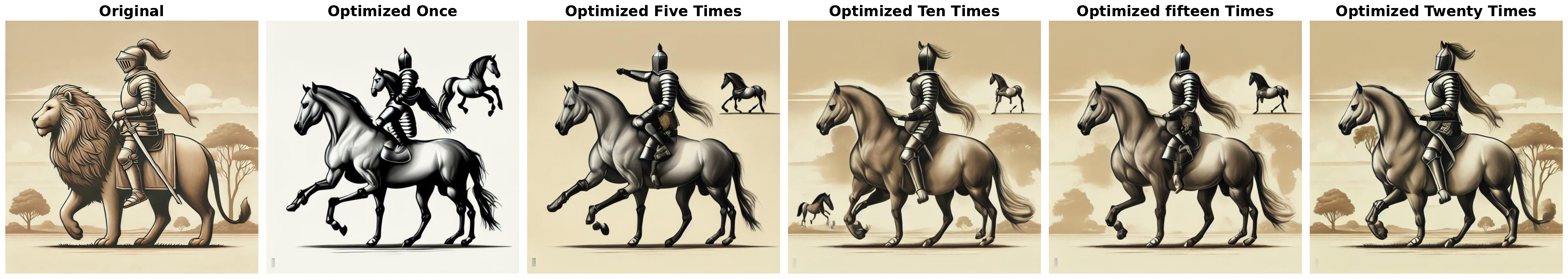}
    \caption{The effect of Bayesian optimization of different numbers of iterations}
    \label{fig:gp}
\end{figure*}

\section{Experiments}

\subsection{Evaluation Protocol}
In evaluating our approach, we carefully considered various metrics commonly used in image processing tasks. While traditional metrics like SSIM and PSNR are widely used, they present significant limitations for instruction-guided image editing evaluation:

\begin{itemize}
    \item \textbf{Pixel-level Comparison:} SSIM and PSNR operate on pixel-level comparisons, which would unfairly penalize intentional edits even when they successfully follow the instructions.
    
    \item \textbf{Semantic Understanding:} These metrics cannot evaluate whether edits align with semantic instructions or distinguish between meaningful changes and random noise.
    
    \item \textbf{Local Edit Evaluation:} For local editing tasks, these metrics would give unreasonably low scores due to pixel changes, even when the edits are successful and appropriate.
\end{itemize}

To address these limitations, we adopt a comprehensive evaluation strategy combining both objective and subjective assessments:

\paragraph{Objective Metrics} We utilize two complementary approaches:
\begin{itemize}
    \item \textbf{CLIP-based Similarity:} To evaluate the semantic alignment between edited images and intended modifications.
    \item \textbf{LLM Scoring:} To assess the nuanced visual content and adherence to editing instructions.
\end{itemize}

\paragraph{Subjective Evaluation} We conducted an anonymous user study approved by our institution's ethics committee. Participants evaluated the edited images based on: (1) faithfulness to the editing requirements, (2) overall visual quality, and (3) preservation of original image context.

\subsection{Experimental Setup}
\paragraph{Baselines Selection}
Most state-of-the-art local editing methods, including ZONE~\cite{li2024zone}, heavily rely on mask priors to constrain the editing regions—either through manual input or derived from operations such as attention map analysis and semantic segmentation. In contrast, our method, to the best of our knowledge, is the first to aim for localized editing without relying on any form of mask guidance. As such, comparing our approach with mask-guided methods would neither be fair nor meaningful.

Instead, we focus our evaluations on methods that, like ours, do not utilize masks or region segmentation in their pipeline:
\begin{itemize}
    \item InstructPix2Pix~\cite{brooks2023instructpix2pix}, which performs editing purely based on text instructions and ranks second to ZONE in their reported experiments.
    \item DALLE-3, which is widely recognized as the current strongest image generation model due to its massive parameter scale.
\end{itemize}
\paragraph{Dataset and Tasks} We constructed a balanced evaluation dataset comprising over 500 images, with the following editing operations:
\begin{itemize}
    \item Adding objects to images
    \item Removing objects from images
    \item Modifying existing objects
\end{itemize}
This diverse set of tasks was selected to comprehensively evaluate our method's versatility and effectiveness across different editing scenarios.

\subsection{Results}

In our experiments, BayesGenie effectively handled the three key image editing tasks, demonstrating robust performance across all scenarios. The results indicate that BayesGenie produces visually consistent images that align closely with user specifications, thanks to the LLM’s multimodal understanding capabilities.

\begin{figure}[ht]
\centering
\includegraphics[width=0.5\textwidth]{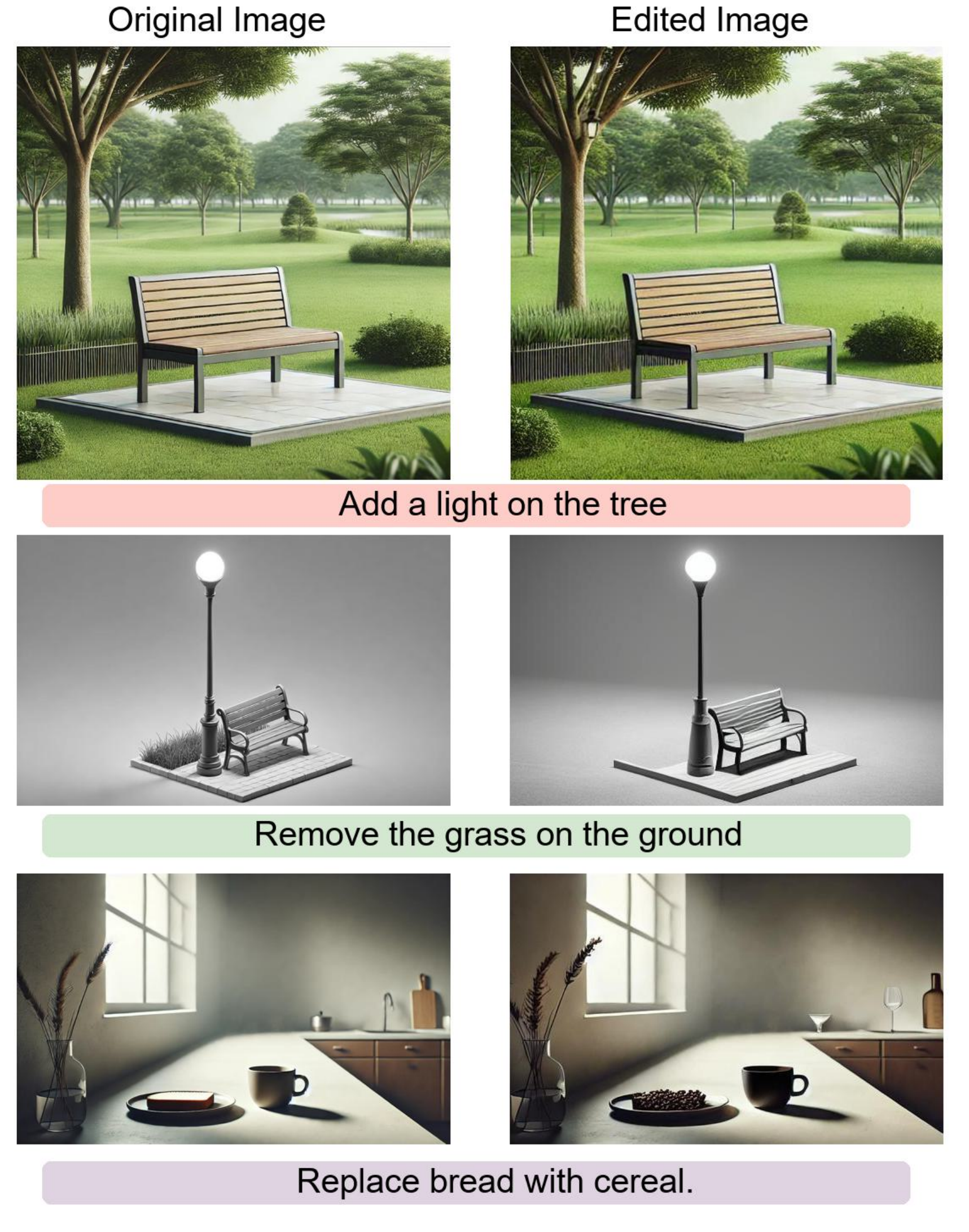}
\caption{Practical application results in different scenarios}
\label{fig:optimise result}
\end{figure}

\paragraph{Qualitative Analysis} 
Figure \ref{fig:gp} shows how Bayesian optimization improves accuracy over iterations, initially with incorrect placement and mismatched background features, but gradually aligning the replacement to match the original image. Figure \ref{fig:optimise result} illustrates our method’s ability to add, remove, and modify elements in a scene while preserving the core features, demonstrating successful addition, removal, and substitution tasks in different visual contexts, showcasing our model's performance in fine-grained image editing.




\begin{figure*}[ht]
\centering
\includegraphics[width=\linewidth]{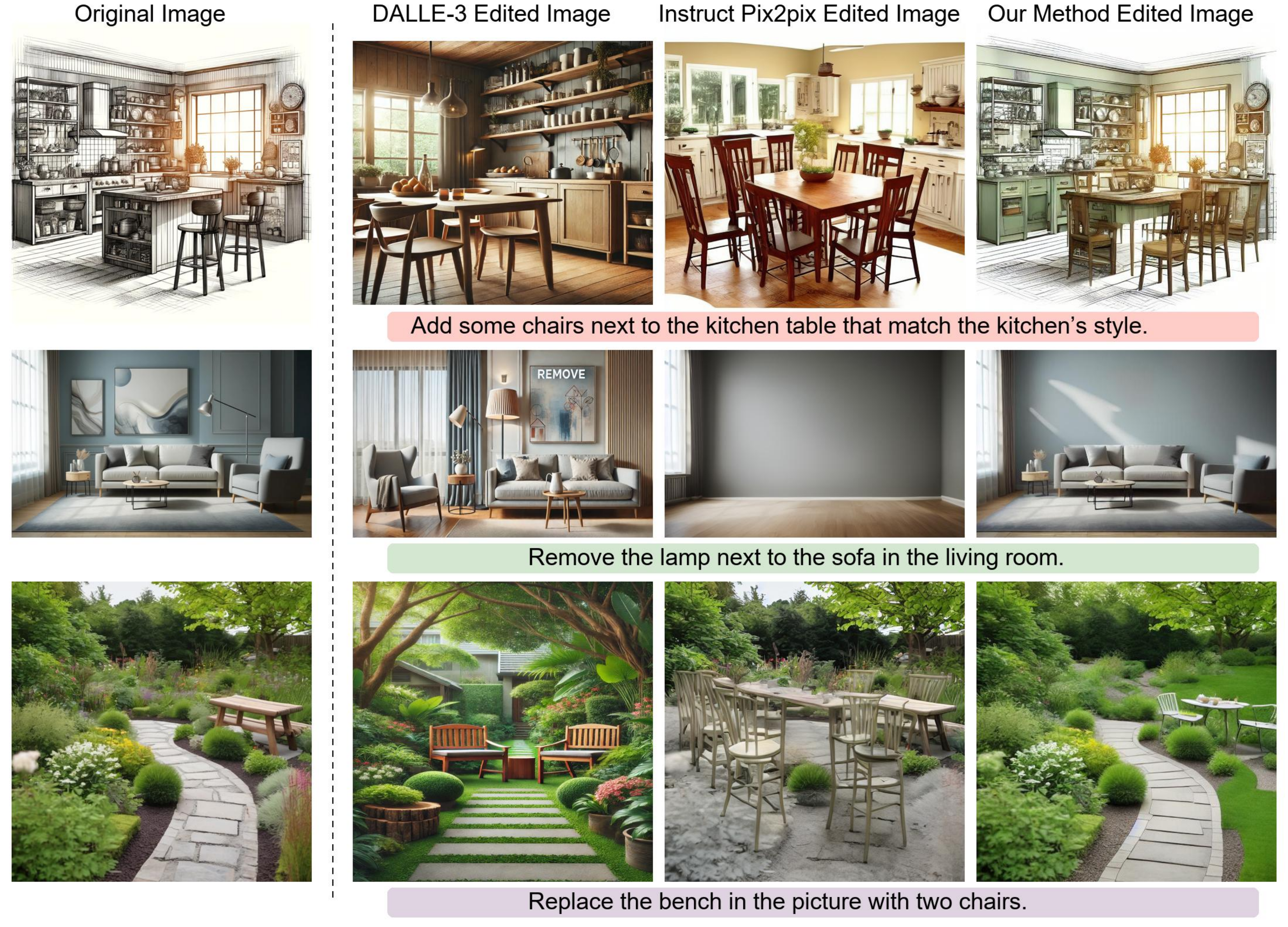}
\caption{Comparison of Image Editing Techniques: DALLE-3, Instruct Pix2pix, and Our method}
\label{fig:compare}
\end{figure*}


\paragraph{Model Comparisons} 
As shown in Figure \ref{fig:compare}, the images generated by DALLE-3 and Instruct Pix2Pix tend to make significant changes to the original images. Although these modifications usually meet the given prompts, they often alter the overall style and content of the image. In contrast, our method preserves the information and characteristics of the original image to a much greater extent, making precise modifications that meet the specified requirements without unnecessary changes.

\begin{figure*}[!htb]
\centering
\includegraphics[width=1.0\linewidth]{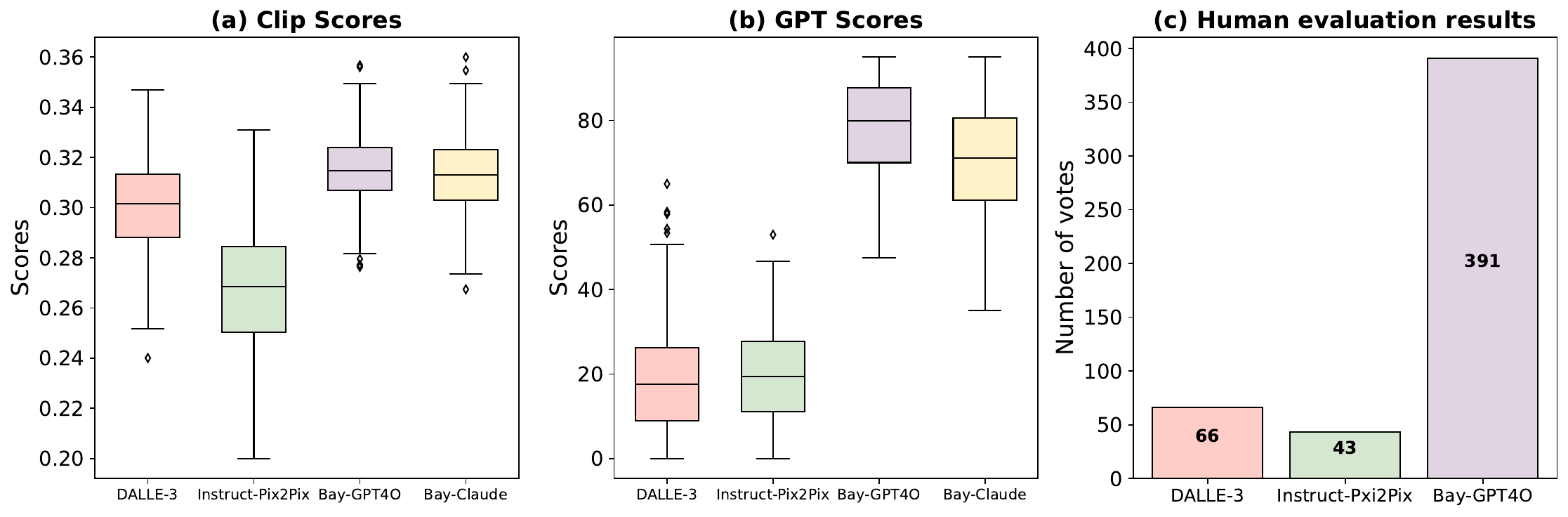}
\caption{Comparison of CLIP scores, GPT scores, and Human evaluation results across different models}
\label{fig:score}
\end{figure*}

\paragraph{Similarity Detection} 
For similarity detection, we primarily relied on the CLIP model's capability to extract features from both images and text, providing a similarity score \cite{hessel2021clipscore}. However, since CLIP evaluates similarities either between images or between images and text, it cannot directly assess the alignment between the modified image and the combined original image plus modification prompt. To address this limitation, we also used ChatGPT to describe the original image, incorporating the modification requirements into the text. We then used this textual description to calculate a GPT score, shown in Figure \ref{fig:score}(b). This approach allowed us to assess the alignment between the combined text and the final output image more effectively.

As shown in the left plot of Figure \ref{fig:score}, our method, including versions using either ChatGPT-4o or Claude 3.5 (referred to as Bay-GPT4o and Bay-Claude), consistently achieved higher CLIP scores than the baseline methods, including Instruct Pix2Pix. This suggests that our method provides greater stability and precision in meeting user expectations. While DALLE-3 performed well, its tendency to make extensive modifications to satisfy the prompts resulted in slightly lower scores.

\paragraph{Human Evaluation:} 
We also conducted a human voting phase to gather subjective evaluations. Participants were asked to select the images they believed best met the editing requirements (see Figure \ref{fig:score}(c)). Instruct Pix2Pix was frequently found to be unsatisfactory due to visible instability and excessive alterations. While DALLE-3 retained some of the original image features, it was selected by fewer participants. In contrast, our method was favored by the majority of participants, demonstrating its higher stability and precision in executing the specified modifications. This result clearly demonstrates the superiority of our approach over the other two methods.

\section{Discussions}
\paragraph{Cost and efficiency}
The algorithm requires relatively low computational resources and costs. The costs for GPT-4o are shown in the table below, with the main expense being the Prompt tokens for Bayesian loop evaluation. The total cost for running the algorithm once to generate a 512x512 image is 0.176 dollar. Additionally, this experiment was conducted on a machine with a single RTX 4080, and each algorithm run takes approximately 2.5 minutes.

In terms of iterations, more iterations represent a finer exploration of the solution space, allowing the algorithm to more precisely converge on an optimal result. Our Bayesian optimization strikes a balance between accuracy and cost by using 20 iterations, as increasing beyond this number has shown diminishing returns in terms of accuracy improvements, while still incurring higher computational costs. This choice ensures that the algorithm remains both computationally efficient and capable of generating high-quality outputs without unnecessary resource consumption.
\begin{table}[htbp]
  \centering
  \caption{Bay-GPT4o Cost}
  \resizebox{\columnwidth}{!}{%
  \begin{tabular}{lcc}
    \toprule
    \textbf{Image size} & \textbf{Generate/Prompt tokens} & \textbf{Cost per case} \\
    \midrule
    512$\times$320 & 5.5k / 24.5k & 0.174 \\
    512$\times$512 & 5.5k / 24.5k & 0.176 \\
    \bottomrule
  \end{tabular}%
  }
  \label{tab:bay-gpt4o-cost} 
\end{table}


\paragraph{Generalizability and Robustness}

In the experiments described above, we demonstrated the robustness of this approach across different LLMs by using GPT-4o and Claude 3.5. Additionally, due to the versatility of the Classifier-Free Guidance (CFG) technique, this method serves as a training-free optimization solution that can be applied to various diffusion models. Future directions may include implementing the text-CFG and image-CFG modules in other diffusion models and testing this method's generalizability.

\section{Conclusion}

Our work introduces BayesGenie, a novel and model-agnostic framework that combines Bayesian Optimization with Large Language Models (LLMs) to enhance the fine-grained image editing process. By leveraging the power of LLMs to generate natural language prompts, BayesGenie simplifies the image creation process, making it more intuitive and accessible. The framework's robust performance across various scenarios highlights its versatility and adaptability, demonstrating its potential for broad applicability in both academic and practical contexts. Ultimately, BayesGenie sets the stage for future advancements in AI-driven content creation, offering a powerful tool for precise and user-friendly image editing.

\section{Ackonwledgement}
This work was supported in part by the Pioneer Centre for AI, DNRF grant number P1.

\bibliography{custom}

\begin{thebibliography}{37}
\providecommand{\natexlab}[1]{#1}

\bibitem[{Alayrac et~al.(2022)Alayrac, Donahue, Luc, Miech, Barr, Hasson, Lenc, Mensch, Millican, Reynolds et~al.}]{alayrac2022flamingo}
Jean-Baptiste Alayrac, Jeff Donahue, Pauline Luc, Antoine Miech, Iain Barr, Yana Hasson, Karel Lenc, Arthur Mensch, Katherine Millican, Malcolm Reynolds, et~al. 2022.
\newblock Flamingo: a visual language model for few-shot learning.
\newblock \emph{Advances in neural information processing systems}, 35:23716--23736.

\bibitem[{Aristodemou et~al.(2025)Aristodemou, Liu, Wang, Kyriakopoulos, Lambotharan, and Wei}]{aristodemou2025maximizing}
Marios Aristodemou, Xiaolan Liu, Yuan Wang, Konstantinos~G Kyriakopoulos, Sangarapillai Lambotharan, and Qingsong Wei. 2025.
\newblock Maximizing uncertainty for federated learning via bayesian optimisation-based model poisoning.
\newblock \emph{IEEE Transactions on Information Forensics and Security}.

\bibitem[{Betker et~al.(2023)Betker, Goh, Jing, Brooks, Wang, Li, Ouyang, Zhuang, Lee, Guo et~al.}]{betker2023improving}
James Betker, Gabriel Goh, Li~Jing, Tim Brooks, Jianfeng Wang, Linjie Li, Long Ouyang, Juntang Zhuang, Joyce Lee, Yufei Guo, et~al. 2023.
\newblock Improving image generation with better captions.
\newblock \emph{Computer Science. https://cdn. openai. com/papers/dall-e-3. pdf}, 2(3):8.

\bibitem[{Boyar and Takeuchi(2024)}]{boyar2024latent}
Onur Boyar and Ichiro Takeuchi. 2024.
\newblock Latent space bayesian optimization with latent data augmentation for enhanced exploration.
\newblock \emph{Neural Computation}, 36(11):2446--2478.

\bibitem[{Brooks et~al.(2023)Brooks, Holynski, and Efros}]{brooks2023instructpix2pix}
Tim Brooks, Aleksander Holynski, and Alexei~A Efros. 2023.
\newblock Instructpix2pix: Learning to follow image editing instructions.
\newblock In \emph{Proceedings of the IEEE/CVF Conference on Computer Vision and Pattern Recognition}, pages 18392--18402.

\bibitem[{Cai et~al.(2024{\natexlab{a}})Cai, Zhao, Du, Liu, and Li}]{cai2024t}
Chengkun Cai, Xu~Zhao, Yucheng Du, Haoliang Liu, and Lei Li. 2024{\natexlab{a}}.
\newblock T\textsuperscript{2} of thoughts: Temperature tree elicits reasoning in large language models.
\newblock \emph{arXiv preprint arXiv:2405.14075}.

\bibitem[{Cai et~al.(2024{\natexlab{b}})Cai, Zhao, Liu, Jiang, Zhang, Wu, Hwang, and Li}]{cai2024role}
Chengkun Cai, Xu~Zhao, Haoliang Liu, Zhongyu Jiang, Tianfang Zhang, Zongkai Wu, Jenq-Neng Hwang, and Lei Li. 2024{\natexlab{b}}.
\newblock The role of deductive and inductive reasoning in large language models.
\newblock \emph{arXiv preprint arXiv:2410.02892}.

\bibitem[{Dhariwal and Nichol(2021)}]{dhariwal2021diffusion}
Prafulla Dhariwal and Alexander Nichol. 2021.
\newblock Diffusion models beat gans on image synthesis.
\newblock \emph{Advances in neural information processing systems}, 34:8780--8794.

\bibitem[{Feng et~al.(2023)Feng, Zhu, Fu, Jampani, Akula, He, Basu, Wang, and Wang}]{NEURIPS2023_3a7f9e48}
Weixi Feng, Wanrong Zhu, Tsu-Jui Fu, Varun Jampani, Arjun Akula, Xuehai He, S~Basu, Xin~Eric Wang, and William~Yang Wang. 2023.
\newblock \href {https://proceedings.neurips.cc/paper_files/paper/2023/file/3a7f9e485845dac27423375c934cb4db-Paper-Conference.pdf} {Layoutgpt: Compositional visual planning and generation with large language models}.
\newblock In \emph{Advances in Neural Information Processing Systems}, volume~36, pages 18225--18250. Curran Associates, Inc.

\bibitem[{Guan et~al.(2025)Guan, Liu, Zhou, Shen, Belongie, Hwang, and Li}]{guan2025learning}
Yunchuan Guan, Yu~Liu, Ke~Zhou, Zhiqi Shen, Serge Belongie, Jenq-Neng Hwang, and Lei Li. 2025.
\newblock Learning to learn weight generation via trajectory diffusion.
\newblock \emph{arXiv preprint arXiv:2502.01117}.

\bibitem[{Hertz et~al.(2022)Hertz, Mokady, Tenenbaum, Aberman, Pritch, and Cohen-Or}]{hertz2022prompt}
Amir Hertz, Ron Mokady, Jay Tenenbaum, Kfir Aberman, Yael Pritch, and Daniel Cohen-Or. 2022.
\newblock Prompt-to-prompt image editing with cross attention control.
\newblock \emph{arXiv preprint arXiv:2208.01626}.

\bibitem[{Hessel et~al.(2021)Hessel, Holtzman, Forbes, Bras, and Choi}]{hessel2021clipscore}
Jack Hessel, Ari Holtzman, Maxwell Forbes, Ronan~Le Bras, and Yejin Choi. 2021.
\newblock Clipscore: A reference-free evaluation metric for image captioning.
\newblock \emph{arXiv preprint arXiv:2104.08718}.

\bibitem[{Isola et~al.(2017)Isola, Zhu, Zhou, and Efros}]{isola2017image}
Phillip Isola, Jun-Yan Zhu, Tinghui Zhou, and Alexei~A Efros. 2017.
\newblock Image-to-image translation with conditional adversarial networks.
\newblock In \emph{Proceedings of the IEEE conference on computer vision and pattern recognition}, pages 1125--1134.

\bibitem[{Jiang et~al.(2024)Jiang, Zhou, Li, Chai, Yang, and Hwang}]{jiang2024back}
Zhongyu Jiang, Zhuoran Zhou, Lei Li, Wenhao Chai, Cheng-Yen Yang, and Jenq-Neng Hwang. 2024.
\newblock Back to optimization: Diffusion-based zero-shot 3d human pose estimation.
\newblock In \emph{Proceedings of the IEEE/CVF Winter Conference on Applications of Computer Vision}, pages 6142--6152.

\bibitem[{Jones et~al.(1998)Jones, Schonlau, and Welch}]{Jones1998}
Donald~R. Jones, Matthias Schonlau, and William~J. Welch. 1998.
\newblock \href {https://doi.org/10.1023/A:1008306431147} {Efficient global optimization of expensive black-box functions}.
\newblock \emph{Journal of Global Optimization}, 13(4):455--492.

\bibitem[{Khan et~al.(2023)Khan, Menouar, and Hamila}]{khan2023crowd}
Muhammad~Asif Khan, Hamid Menouar, and Ridha Hamila. 2023.
\newblock Crowd counting in harsh weather using image denoising with pix2pix gans.
\newblock In \emph{2023 38th International Conference on Image and Vision Computing New Zealand (IVCNZ)}, pages 1--6. IEEE.

\bibitem[{Kushner(1964)}]{10.1115/1.3653121}
H.~J. Kushner. 1964.
\newblock \href {https://doi.org/10.1115/1.3653121} {{A New Method of Locating the Maximum Point of an Arbitrary Multipeak Curve in the Presence of Noise}}.
\newblock \emph{Journal of Basic Engineering}, 86(1):97--106.

\bibitem[{Li et~al.(2024{\natexlab{a}})Li, Jia, Jianhao, Jiang, Zhou, Dai, Zhang, Zongkai, and Hwang}]{li2024human}
Lei Li, Sen Jia, Wang Jianhao, Zhongyu Jiang, Feng Zhou, Ju~Dai, Tianfang Zhang, Wu~Zongkai, and Jenq-Neng Hwang. 2024{\natexlab{a}}.
\newblock Human motion instruction tuning.
\newblock \emph{arXiv preprint arXiv:2411.16805}.

\bibitem[{Li et~al.(2024{\natexlab{b}})Li, Zeng, Feng, Gao, Liu, Liu, Li, Tang, Hu, Liu et~al.}]{li2024zone}
Shanglin Li, Bohan Zeng, Yutang Feng, Sicheng Gao, Xiuhui Liu, Jiaming Liu, Lin Li, Xu~Tang, Yao Hu, Jianzhuang Liu, et~al. 2024{\natexlab{b}}.
\newblock Zone: Zero-shot instruction-guided local editing.
\newblock In \emph{Proceedings of the IEEE/CVF Conference on Computer Vision and Pattern Recognition}, pages 6254--6263.

\bibitem[{Li et~al.(2024{\natexlab{c}})Li, Guan, Wei, Zhou, Zhang, and Xu}]{li2024mapping}
Zhenglin Li, Bo~Guan, Yuanzhou Wei, Yiming Zhou, Jingyu Zhang, and Jinxin Xu. 2024{\natexlab{c}}.
\newblock Mapping new realities: Ground truth image creation with pix2pix image-to-image translation.
\newblock \emph{arXiv preprint arXiv:2404.19265}.

\bibitem[{Liu et~al.(2024)Liu, Chen, Jia, Shi, Jiang, Jin, Zongkai, Hwang, and Li}]{liu2024graph}
Libin Liu, Shen Chen, Sen Jia, Jingzhe Shi, Zhongyu Jiang, Can Jin, Wu~Zongkai, Jenq-Neng Hwang, and Lei Li. 2024.
\newblock Graph canvas for controllable 3d scene generation.
\newblock \emph{arXiv preprint arXiv:2412.00091}.

\bibitem[{M~Shetty K~Raghavendra(2022)}]{m2022transfer}
Prasad Narasimha~Sarappadi M~Shetty K~Raghavendra. 2022.
\newblock Transfer learning with pix2pix gan for generating realistic photographs from viewed sketch arts.
\newblock \emph{Journal of Southwest Jiaotong University}, 57(4).

\bibitem[{OpenAI(2023)}]{openai2023}
OpenAI. 2023.
\newblock \href {https://www.openai.com/research/gpt-4} {Gpt-4 technical report}.

\bibitem[{Patashnik et~al.(2021)Patashnik, Wu, Shechtman, Cohen-Or, and Lischinski}]{patashnik2021styleclip}
Or~Patashnik, Zongze Wu, Eli Shechtman, Daniel Cohen-Or, and Dani Lischinski. 2021.
\newblock Styleclip: Text-driven manipulation of stylegan imagery.
\newblock In \emph{Proceedings of the IEEE/CVF international conference on computer vision}, pages 2085--2094.

\bibitem[{Radford et~al.(2021)Radford, Kim, Hallacy, Ramesh, Goh, Agarwal, Sastry, Askell, Mishkin, Clark et~al.}]{radford2021learning}
Alec Radford, Jong~Wook Kim, Chris Hallacy, Aditya Ramesh, Gabriel Goh, Sandhini Agarwal, Girish Sastry, Amanda Askell, Pamela Mishkin, Jack Clark, et~al. 2021.
\newblock Learning transferable visual models from natural language supervision.
\newblock In \emph{International conference on machine learning}, pages 8748--8763. PMLR.

\bibitem[{Ramesh et~al.(2022)Ramesh, Dhariwal, Nichol, Chu, and Chen}]{ramesh2022hierarchical}
Aditya Ramesh, Prafulla Dhariwal, Alex Nichol, Casey Chu, and Mark Chen. 2022.
\newblock Hierarchical text-conditional image generation with clip latents.
\newblock \emph{arXiv preprint arXiv:2204.06125}, 1(2):3.

\bibitem[{Rombach et~al.(2022)Rombach, Blattmann, Lorenz, Esser, and Ommer}]{rombach2022high}
Robin Rombach, Andreas Blattmann, Dominik Lorenz, Patrick Esser, and Bj{\"o}rn Ommer. 2022.
\newblock High-resolution image synthesis with latent diffusion models.
\newblock In \emph{Proceedings of the IEEE/CVF conference on computer vision and pattern recognition}, pages 10684--10695.

\bibitem[{Saharia et~al.(2022)Saharia, Ho, Chan, Salimans, Fleet, and Norouzi}]{saharia2022image}
Chitwan Saharia, Jonathan Ho, William Chan, Tim Salimans, David~J Fleet, and Mohammad Norouzi. 2022.
\newblock Image super-resolution via iterative refinement.
\newblock \emph{IEEE transactions on pattern analysis and machine intelligence}, 45(4):4713--4726.

\bibitem[{Shi et~al.(2024)Shi, Li, Ma, Yang, Ma, and Li}]{shi2024chops}
Jingzhe Shi, Jialuo Li, Qinwei Ma, Zaiwen Yang, Huan Ma, and Lei Li. 2024.
\newblock Chops: Chat with customer profile systems for customer service with llms.
\newblock \emph{arXiv preprint arXiv:2404.01343}.

\bibitem[{Shi et~al.(2025)Shi, Ma, Liu, Zhao, Hwang, Belongie, and Li}]{shi2025explaining}
Jingzhe Shi, Qinwei Ma, Hongyi Liu, Hang Zhao, Jeng-Neng Hwang, Serge Belongie, and Lei Li. 2025.
\newblock Explaining context length scaling and bounds for language models.
\newblock \emph{arXiv preprint arXiv:2502.01481}.

\bibitem[{Snoek et~al.(2012)Snoek, Larochelle, and Adams}]{snoek2012practical}
Jasper Snoek, Hugo Larochelle, and Ryan~P Adams. 2012.
\newblock Practical bayesian optimization of machine learning algorithms.
\newblock \emph{Advances in neural information processing systems}, 25.

\bibitem[{Van Den~Oord et~al.(2016)Van Den~Oord, Kalchbrenner, and Kavukcuoglu}]{van2016pixel}
A{\"a}ron Van Den~Oord, Nal Kalchbrenner, and Koray Kavukcuoglu. 2016.
\newblock Pixel recurrent neural networks.
\newblock In \emph{International conference on machine learning}, pages 1747--1756. PMLR.

\bibitem[{Wang et~al.(2024)Wang, Dahl, Swersky, Lee, Nado, Gilmer, Snoek, and Ghahramani}]{JMLR:v25:23-0269}
Zi~Wang, George~E. Dahl, Kevin Swersky, Chansoo Lee, Zachary Nado, Justin Gilmer, Jasper Snoek, and Zoubin Ghahramani. 2024.
\newblock \href {http://jmlr.org/papers/v25/23-0269.html} {Pre-trained gaussian processes for bayesian optimization}.
\newblock \emph{Journal of Machine Learning Research}, 25(212):1--83.

\bibitem[{Yao et~al.(2024)Yao, Li, Zhou, Liu, Jiang, Wang, Zheng, Zou, and Li}]{yao2024car}
Ziyu Yao, Jialin Li, Yifeng Zhou, Yong Liu, Xi~Jiang, Chengjie Wang, Feng Zheng, Yuexian Zou, and Lei Li. 2024.
\newblock Car: Controllable autoregressive modeling for visual generation.
\newblock \emph{arXiv preprint arXiv:2410.04671}.

\bibitem[{Zhang et~al.(2023)Zhang, Mo, Chen, Sun, and Su}]{zhang2023magicbrush}
Kai Zhang, Lingbo Mo, Wenhu Chen, Huan Sun, and Yu~Su. 2023.
\newblock Magicbrush: A manually annotated dataset for instruction-guided image editing.
\newblock \emph{Advances in Neural Information Processing Systems}, 36:31428--31449.

\bibitem[{Zhao et~al.(2024)Zhao, Ma, Chen, Si, Wu, An, Yu, Zhang, Li, and Chang}]{zhao2024ultraedit}
Haozhe Zhao, Xiaojian~Shawn Ma, Liang Chen, Shuzheng Si, Rujie Wu, Kaikai An, Peiyu Yu, Minjia Zhang, Qing Li, and Baobao Chang. 2024.
\newblock Ultraedit: Instruction-based fine-grained image editing at scale.
\newblock \emph{Advances in Neural Information Processing Systems}, 37:3058--3093.

\bibitem[{Zhu et~al.(2017)Zhu, Park, Isola, and Efros}]{zhu2017unpaired}
Jun-Yan Zhu, Taesung Park, Phillip Isola, and Alexei~A Efros. 2017.
\newblock Unpaired image-to-image translation using cycle-consistent adversarial networks.
\newblock In \emph{Proceedings of the IEEE international conference on computer vision}, pages 2223--2232.

\end{thebibliography}

\appendix
\section{Examples of Prompt Generation}
In this section, we present examples of input images alongside the corresponding requirements and GPT-4 generated prompts. These examples illustrate how the Large Language Model (LLM) interprets user instructions and generates detailed prompts that guide the image modification process.
\begin{figure}[h]
    \centering
    \includegraphics[width=0.5\textwidth,height=0.25\textheight]{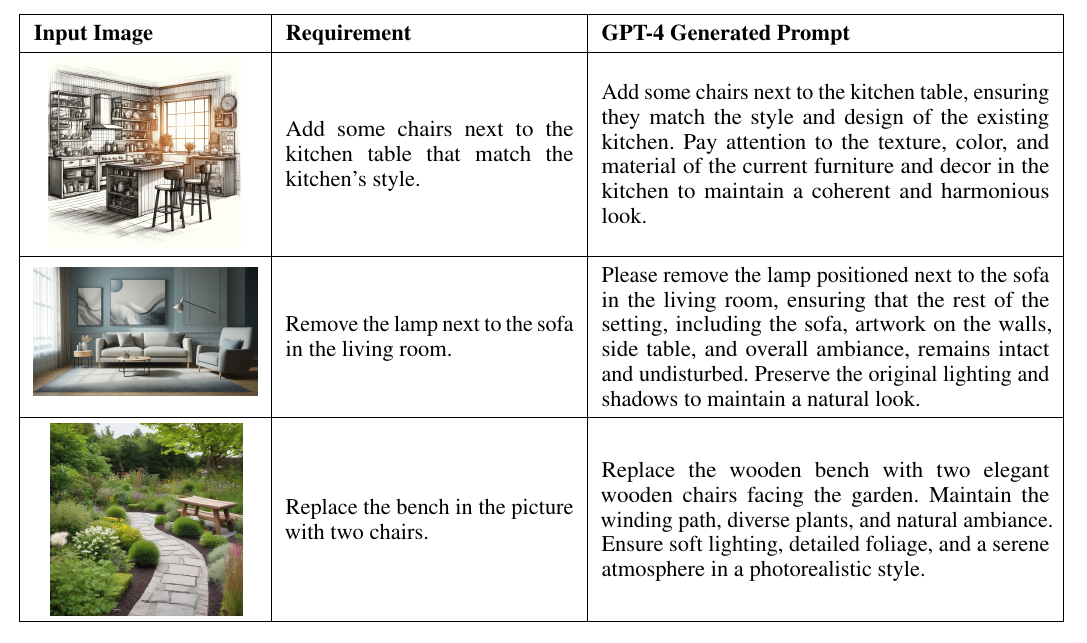}
    \caption{Examples of Input Images, Requirements, and GPT-4 Generated Prompts}
    \label{fig:prompt_example}
\end{figure}

\section{Scoring Evaluation Prompt}
This appendix provides the specific prompt used for evaluating the modified images generated by the model. The prompt is designed to ensure that the generated images adhere to the specified requirements, while also maintaining the overall integrity and coherence of the original image. The evaluation prompt outlines the criteria for scoring the generated image, which includes assessing the degree of alteration and its alignment with the original content. 
\begin{quote}
        {
        The following is a requirement for modifying an image:
        
        Below are two images: the original image and the generated image after modification.

        The first one is the original Image.
        
        The second one is the generated Image.

        Please evaluate whether the generated image meets the requirement. Provide a score from 0 to 100 based on the following criteria:
        \newline
        1. If the generated image is altered too much compared to the original image, give a low score.
        \newline
        2. If the generated image is altered too little, give a low score.
        \newline
        3. If the generated image meets the requirement well, give a high score.
        \newline
        The return should begin with: The score is:
        \newline
        Ensure the scores follow a normal distribution, with the majority of scores being around the middle range, and only exceptional cases scoring very low or very high. Also, provide a brief explanation for the score.
        }
        \end{quote}

\section{Comparison with State-of-the-Art Models}
To further validate the effectiveness and generalizability of our proposed BayesGenie framework, we conducted additional experiments comparing it with several recent state-of-the-art image editing models, including MagicBrush \cite{zhang2023magicbrush} and UltraEdit \cite{zhao2024ultraedit}. The evaluation metrics include CLIPScore and GPT-4o-based instruction consistency scores. Results are summarized in Table~\ref{tab:sota_comparison}.

\begin{table}[h]
\centering
\caption{Performance comparison with state-of-the-art models.}
\label{tab:sota_comparison}
\resizebox{\columnwidth}{!}{%
\begin{tabular}{lcc}
\toprule
\textbf{Model} & \textbf{CLIPScore} & \textbf{GPT-4o Score} \\
\midrule
InstructPix2Pix & 0.2712 & 23.5 \\
InstructPix2Pix + BayesGenie & 0.3180 & 78.6 \\
MagicBrush & 0.3078 & 53.7 \\
UltraEdit & 0.3302 & 62.6 \\
UltraEdit + BayesGenie & \textbf{0.3524} & \textbf{85.3} \\
\bottomrule
\end{tabular}
}
\end{table}

To enable a fair and consistent comparison, we integrated \textbf{BayesGenie} into both the InstructPix2Pix and UltraEdit pipelines. InstructPix2Pix and UltraEdit both support classifier-free guidance (CFG), which BayesGenie optimizes during inference. For UltraEdit, which typically uses spatial masks, we covered the entire image as the editable region to align it with the mask-free nature of BayesGenie.

We evaluated each model on 21 diverse editing tasks (including addition, removal, and modification of objects), with each task repeated five times to report averaged scores. It is worth noting that both MagicBrush and InstructPix2Pix utilize Stable Diffusion 1.5 as their backbone, whereas UltraEdit is based on Stable Diffusion 3, which partially accounts for some of the performance differences observed.

The results highlight two core strengths of the \textbf{BayesGenie} framework:

\begin{enumerate}
    \item \textbf{Complementary Nature:} \textbf{BayesGenie} significantly improves the performance of existing models such as InstructPix2Pix and UltraEdit, not by competing with them, but by acting as an effective enhancement module.
    
    \item \textbf{Universal Applicability:} \textbf{BayesGenie} demonstrates strong performance across a wide spectrum of architectures, from general-purpose models like InstructPix2Pix to highly specialized systems like UltraEdit. This underscores its versatility and model-agnostic design.
\end{enumerate}

\section{Ethics and Participant Consent}
This study was conducted following the ethical guidelines established by our institution’s Research Ethics Committee, all participants in the study provided informed consent before participating. Participants were fully informed about the purpose of the study, which aimed to evaluate AI-generated image modifications. Participation was voluntary, with the right to withdraw at any point before submitting the survey responses. No personally identifiable information was collected, ensuring complete anonymity. Anonymised data was used for academic publications and presentations, with participants given the option to consent to future use in ethically approved research. Participants filled out the survey via social media platforms and were offered a chance to win a £10 Amazon gift card as an incentive for their participation.

\end{document}